# Towards an Accurate and Effective Robot Vision

## The Problem of Topological Localization for Mobile Robots

by

Emanuela Boroş

B.Sc., Alexandru Ioan Cuza University, 2010
B.Sc., George Enescu Art University, 2010

MASTER OF SOFTWARE ENGINEERING

ALEXANDRU IOAN CUZA UNIVERSITY

FACULTY OF COMPUTER SCIENCE

Signature of Author .................................................................
Faculty of Computer Science
July 2012

Certified by .................................................................
Adrian Iftene
Lecturer PhD
Thesis Supervisor

# Towards an Accurate and Effective Robot Vision

## The Problem of Topological Localization for Mobile Robots

by

Emanuela Boroş

# Abstract

**Topological localization** is a fundamental problem in mobile robotics. Most mobile robots must be able to self-locate in their environment in order to accomplish their tasks. Robot visual localization and place recognition are not easy tasks, and this is mainly due to the perceptive ambiguity of acquired data and the sensibility to noise and illumination variations of real world environments. Being able to identify the robot's location constitutes the foundations on which high-level reasoning processes conducted by a robot will build up. In order to help reduce this gap, this work addresses the problem of topological localization of a robot in an office environment.

We approach the task of topological localization without using a temporal continuity of the sequences of images. The provided information about the environment is contained in images taken with a perspective color camera mounted on a robot platform. The state of the art methods were selected: Color Histograms, SIFT (Scale Invariant Feature Transform), ASIFT and RGB-SIFT (more compact variants for SIFT), Bag-of-Visual-Words strategy inspired from the text retrieval community.

The main contributions of this work are quantifiable examinations of a wide variety of different features, different distance measures. The experiments consist an extension of our previous work. We focused on finding the optimal set of features and and a deepend analysis was carried out, the characteristics of different features were analyzed using widely known performance measures and graphical views of the overall performances.

After evaluating these methods in challenging datasets, different conclusions were drawn with leading advantages in using proper configurations of visual-based appearance descriptors, similarity measures and classifiers. The quality of the acquired configurations is also tested in the localization stage by means of location recognition in the task Robot Vision, by participating at the International Evaluation Campaign ImageCLEF: given a new sequence of images, the most likely location where that view came had to be determined.

Future work consists in combining different hierarhical models, ranking methods and features in order to create a more robust localization recognition classifier while reducing the training and utilization time and to avoid the curse of dimensionality. The long term goal of this project is to develop integrated, stand alone capabilities for real-time topological localization in varying illumination conditions and over longer routes.

**Keywords:** *Visual Place Classification, Robot Topological Localization, Visual Feature Detectors, Visual Feature Descriptors*

**Emanuela Boroş**             emanuela.boros@info.uaic.ro

Thesis Supervisor: Adrian Iftene
Title: Lecturer PhD

# Chapter 1

# Table of Contents

# Chapter 2

# List of Tables

# Chapter 3

# List of Figures

# Preface

- 2012

    - *To appear:* **Boroş, E.** Towards an Accurate Topological Localization using a Bag-of-SIFT-Visual-Words Model. In *IEEE 9th International Conference on Intelligent Computer Communication and Processing*, Cluj-Napoca, România, August 30-September 1, 2012.
    - *To appear:* **Boroş, E.**, Gînscă, A. L., Iftene, A. UAIC: Participation in ImageCLEF 2012 Robot Vision Task. In *Proceedings of the CLEF 2012 Workshop*. Rome, Italy, September 17-September 20, 2012.

- 2011

    - **Boroş, E.**, Gînscă, A. L., Iftene, A. UAIC's Participation at Wikipedia Retrieval @ ImageCLEF 2011. In *Notebook Paper for the CLEF 2011 LABs Workshop*, ISBN 978-88- 904810-1-7, ISSN 2038-4726, Amsterdam, Netherlands, September 19-22, 2011.
    - Gînscă, A. L., **Boroş, E.**, Metzak A., Florea A. PETI Patient Eye Tracking Interface. In *Proceedings of the 8th National Conference on Human-Computer Interaction RoCHI* 2011, ISSN 1843-4460, Pages 107-110. Bucharest, România, September 8-9, 2011.
    - **RoCHI** Workshop

- 2010

    - **Boroş, E.**, Roşca, G., Iftene, A. Using SIFT Method for Global Topological Localization for Indoor Environments. In *C. Peters et al. (Eds.): CLEF 2009, LNCS 6242, Part II (Multilingual Information Access Evaluation Vol. II Multimedia Experiments)*. Pages 277-282. ISBN: 978-3-642-15750-9. Springer, Heidelberg, 2010.
    - **BringITon** Workshop

- 2009

    - **Boroş, E.**, Roşca, G., Iftene, A. UAIC: Participation in ImageCLEF 2009 Robot Vision Task. In *Proceedings of the CLEF 2009 Workshop*. Corfu, Greece, September 30-October 2, 2009.

# Declaration of originality and copyright compliance

I hereby declare that the license paper titled *Towards an Accurate and Effective Robot Vision* is written by me and never has been presented at another university or institution of higher education in the country or abroad.

I also declare that all sources used, including those downloaded from the Internet, or indicated in the paper with the rules to avoid plagiarism:

- all fragments of text reproduced exactly, even in his own translation from another language, are written in quotes and accurate reference of the source;

- reformulation in their own words of written texts by other authors has accurate reference;

- source code, images, ... taken from open source projects or other sources are used with copyright compliance and have accurate references;

- summarizing the ideas of other authors state the precise reference to the original.

Iaşi, **Graduate:** Emanuela Boroş

# Declaration of consent

I hereby declare that the master's thesis entitled *Towards an Accurate and Effective Robot Vision*, source code and other content (graphics, multimedia, test data, etc.) accompanying this paper to be used in the Faculty of Computer Science.

I also agree that the Faculty of Computer Science at the Alexandru Ioan Cuza University to use, modify, reproduce and distribute non-commercial purposes the computer programs, in executable or source code format, made by me in this master's thesis.

Iași,
**Dean:** Lecturer, PhD. Adrian Iftene          **Graduate:** Emanuela Boroş

# Agreement on copyright ownership

Faculty of Computer Science agrees that the copyright in computer programs, executable and source code format belong to the author, Emanuela Boroş.

A joint agreement is required for the following reasons:

Iaşi, **Graduate:** Emanuela Boroş

# Chapter 4

# Introduction

**Computer Vision** = *The science and engineering discipline concerned with making inferences about the external world, given one or more of its images.*
**Topological localization** = *The problem of recognizing environmental settings by making inferences from information that computer vision provides.*

Computer vision basically implies the process where the information is reliably extracted from the images and through assumptions and knowledge processing that are applied at different levels and after that, meaningful descriptions of objects or scenes are composed. These descriptions have the means of feature vectors and are a prerequisite for image recognition, modeling and classification.

We are interested in how an image is perceived in computer vision what properties does it have, how can we use these properties in our advantage for constructing an accurate image classifier of images, accordingly to D. Ballard [2] who said that *The challenge of computer vision is one of explicitness. Exactly what information about scenes can be extracted from an image using only very basic assumptions about physics and optics? Explicitly, what computations must be performed? Then, at what stage must domain dependent, prior knowledge about the world be incorporated into the understanding process?*

According to R. Haralick in [26] *Computer Vision is the combination of image processing, pattern recognition, and artificial intelligence technologies which focuses on the computer analysis of one or more images, taken with a single-band/multiband sensor, or taken in time sequence. The analysis recognizes and locates the position and orientation, and provides a sufficiently detailed symbolic description or recognition of those imaged objects deemed to be of interest in the three-dimensional environment. The Computer Vision process often uses geometric modeling and complex knowledge representations in an expectation- or model-based matching or searching methodology.*

This thesis covers a specific subject in computer science which refers to topological localization. This discussion needs an introduction in robot vision field and also a suite of benefits that this vision technology brings in which are included automated identification and navigation. The robot vision paradigm is described as being a trend of research and development which is developing rapidly. Mainly, robot vision is used for part identification and navigation. The present state of robotics entails that industrial robots carry out recurring simple tasks in a fast, accurate and reliable manner. This is a typical case in applications of series production where the environment is customized in relation to a fixed location and volume occupied by the robot and/or the robot is built such that certain spatial relations with a fixed environment are kept. Task-relevant effector trajectories must be planned perfectly during an offline phase and unexpected events must not occur during the subsequent online phase. The benefits of sophisticated vision technology include savings, improved quality, reliability, safety and productivity. The aims of this field are automatic planning systems are available which generate robot programs for assembly or disassemble tasks, for example sequences of movement steps of the effector for assembling complex objects out of components. To create more then a system that assumes the initial state and the desired state of the task as known and to be able to control their actions in a way that non-deterministic incidents with the robot or in the environment can be perceived and acted upon or to be able to adapt to the new environment or objects in contact, the perception of unusual events in the environment.

Robot vision applications generally deal with finding a part and orienting it for robotic handling or inspection before an action is performed. Sometimes vision guided robots can replace multiple mechanical tools with a single robot station. A combination of vision algorithms, calibration, temperature software, and cameras provide the vision ability. Calibration of robot vision system is very application dependent. They can range from a simple guidance application to a more complex application that uses data from multiple sensors. Algorithms are consistently improving, allowing for sophisticated detection. Many robots are now available with collision detection, allowing them to work alongside other robots without the fear of a major collision. They simply stop moving momentarily if they detect another object in their motion path.

The main approach to these problems is to perceive the environment continually and make use of the reconstructed spatial relations between robot and target objects. In addition to the close-range sensors one substantially needs long-range perception devices such as video, laser, infrared, and ultrasonic cameras. The specific limitations and constraints, which are inherent in the different perception devices, can be compensated by a fusion of the different image processing modalities.

Image analysis is the basic means for the primary goal of reconstructing the robot-object relations. To be useful, the image analysis system must provide relevant information in amounts of time to construct an adequate image understanding system which is an important issue in this field. Also, it is hard to impossible to proof the correctness of reconstructed scene information and this is the reason why the development and application of camera-equipped robots is restricted and oriented to research institutes.

There are many cases in which robot vision cannot overcome the problem of analyzing the environment, for example the natural scenes where their dynamical and multiple possibilities of changing cannot be presumed and image analysis becomes inefficient and hard to compute. The camera-equipped robots should be able to work in completely different environments and carrying out new categories of tasks. Examples of such tasks include supporting disabled persons at home, cleaning rooms in office buildings, doing work in hazardous environments, automatic modeling of real objects or scenes, etc. These tasks have in common that objects or scenes must be detected in the images and reconstructed with greater or lesser degree of detail.

In practical applications, these robot systems are lacking correct and goal-oriented image-motor mappings. This finding can be traced back to the lack of correctness of image processing, feature extraction, and reconstructed scene information. Our most important goal is find a new concept for the development and evaluation of image analysis methods and image classifying towards an accurate and effective robot localization.

## 4.1 Overview

This work introduces a practical methodology for developing a system of scene recognition analyzing the feedback from a camera-equipped robot systems. We addressed the problem of topological localization of a mobile robot using visual information determining the topological location of a robot based on images acquired with a perspective camera mounted on a robot platform.

The training data consists of an image sequence recorded in a nine room subsection of an indoor environment under varying illumination conditions and at a given time. The main challenge is to build a system capable of answering the question *Where are you?* (*I'm in the Corridor, in the LoungeArea etc.*) when presented with a test sequence containing images acquired under different illumination settings or in additional rooms that were not imaged in the training sequence or with different furniture and different people wondering in the rooms.

The system should assign each test image to one of the office section that were present in the training sequence or indicate that the image comes from a room that was not included during training. Moreover, the system can refrain from making a decision, in the case of lack of confidence and declare that image from a room that it doesn't know. An important task that is approached is to provide information about the location of the robot separately for each test image without taking in consideration the continuity of the robot's route, maybe even testing on scrambled sequences. This corresponds basically to the problem of global topological localization.

In this section we present an overview of this task that computer vision has been applied to, with a great degree of success. We begin with a generic architecture that many computer vision systems follow, and then discuss a few of the tasks within the context of this architecture. The goal of this chapter is to present the basic architecture of the needed system that will serve as the building blocks for the methods developed in this thesis.

For the purposes of this thesis, data acquisition process consists from images taken in an office environment, a data set of 7112 images and 9 categories, varying by number per category, from 264 *(Printer Room)* to 1651 *(Student Office)*. The data is a set of digital images containing intensities for each pixel as well as color information in some cases, as there are images without these properties being taken in dark corners or under low illumination settings. An overview over the main subjects that should be addressed implies, first of all, an image function that becomes the most important abstraction of an image, then, a geometrical model describes the projection of three dimensions into two, next, the properties of imaging geometry and light measurement that affect the image analysis, a spatial frequency model that describes spacial variation of the image, a color model with different spectral measurements and a final model that brings us to obtaining discrete samples.

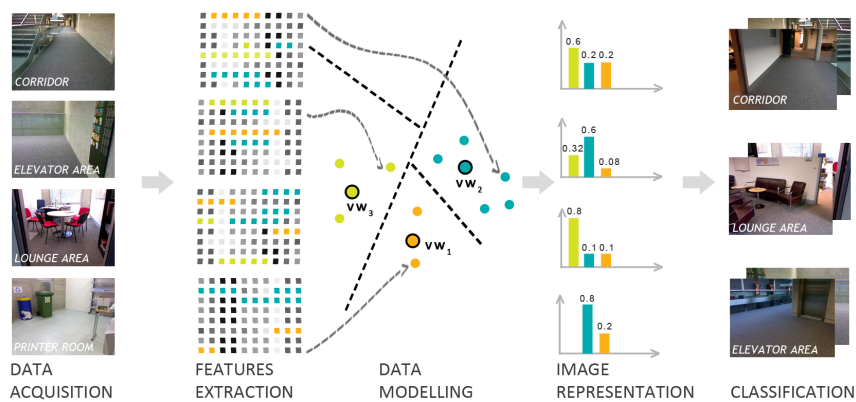

Figure 4.1: Basic Architecture

In a typical computer vision application, after acquiring the images, a form of data reduction is performed to help in the subsequent analysis. This in itself is biologically inspired: the retina has about 120 million rods and 5 million cones, but the total number of nerve fibers leaving the eye is only about 1 million. The chosen form of compression is applied very early in image processing, in the pre-processing step is to express images in a format that is better for computing features. One of the main reasons that can occur in this process is that images can have high resolutions and then the data extracted from them can become redundant for the next steps and also can mean a large computing effort. The most common approaches are down sampling the images at resolutions acceptable or converting color images to gray scale or even filtering for noise reduction to lower high frequency content.

Following preprocessing of the image, Figure 4.1 shows the next step as being the features extraction process which basically means a type of image segmentation into units or an overview of the characteristics of an image that will be used in further stages. Segmentation has two predominant categories: image regions known as *patches* and interest points. The first type groups pixels into regions that cover the entire image, as slicing an image in different grids. These regions might be disjoint *over-segmentation* of the image, where each pixel belongs to one region that attempts to preserve some coherence between the pixels in the region, or they might be overlapping patches of the image, typically square patches laid out in a grid segmentation.

The other type of segmentation attempts to produce a sparse representation of the image, based on interest points, or places in the image where it is likely that something important is happening. Many interest operators have been proposed, generally based on the idea of finding either corners [43], areas of high gradients [38], or other *salient* regions in the image. The extracted keypoints often include a scale or orientation, which allows us to determine which pixels contribute to this point. The global features imply an aggregate of histograms, shape, texture. Such features are important because they produce very compact representations of images, where each image corresponds to a point in a high dimensional feature space.

The next step consists in attaching a set of features to each *patch*. This set of image features often takes the form of a vector of real numbers computed from the pixels in the neighborhood of the segmentation unit. In the case of large regions, this might include the mean and standard deviations of several quantities across the pixels in the region. For interest points, the feature vector is usually computed from a small region around the keypoint, based on the scale and orientation of the point. The well-known scale-invariant feature transform (SIFT) [8, 38, 39] feature is a good example of this type of feature.

To efficiently handle these descriptors, the key idea is in the next step shown in Figure 4.1 reffered as data modeling and consists in the quantization of each extracted keypoint into one of *visual words*, and then represent each image by a distribution of the visual words. *Visual words* are *iconic* image patches or keypoints. This vector quantization procedure allows the representation of each image by a histogram of the *visual words*, which is often referred to as the *bag-of-visual-words* representation.

The final step of the computer vision pipeline is the classification of new test data using the obtained training data in order to discriminate vectors corresponding to positive and negative training images.

We are channeling our proposal on the ability of easily and accurately distinguish the sections of an office environment excluding unwanted information which may confuse the object recognition method, increasing the capability for a detection method to be able to generalize and recognize previously unseen instances, dealing with only partial information of the image, taking into account motion blur situations, repetitive patterns which cause problems in methods that have a data association stage, data scaling so the descriptor distributions won't become different, concentrating on the informative parts of the image the size of the redundant input data is significantly reduced and, on the second place, the method is insensitive to small pixel intensity variations due to noise in the pixels, as well as small changes in point of view, scale or illumination and we want to measure the frame-rate at which comparable implementations of each method can work.

The remainder of this document is organized as follows: Chapter 5 presents the underlying principles of image analysis, with detailed descriptions of global and local feature descriptors. Chapter 6 gives a survey of matching distance between images, different distance dissimilarity measures. In Chapter 7 we introduce the *bag-of-visual-words* data modeling technique, strictly applied in image recognition and classification field. Chapter 8 presents and compares the effectiveness of our approaches to topological localization along with experiments performed with different features, matching distances and modeling differences. Finally, we conclude with a discussion (Chapter 9) about the remaining aspects that are still unsolved problems, we conclude this work in Chapter 10 and propose further research.

## 4.2 Related Work

Nowadays, an increasing number applications need robot localization as a prerequisite and research has been very active in this field.

In this paper, we present an approach to vision-based mobile robot localization that uses a single perspective camera taken within an office environment. The robot should be able to answer the question *where are you?* when presented with a test sequence representing a room category seen during training [40, 53, 57]. We analyze the problem without taking in consideration the use of the temporal continuity of the sequences of images. In this paper the comparison is carried out for different matching approaches used for the creation of visual words vocabularies. We perform a more exhaustive evaluation and introduce new analysis statistic between quantization solutions.

Many approaches during last years have been developed using different methods for robotic topological localization such as topological map building which makes good use of temporal continuity [63], simultaneous localization and mapping [12], using Monte-Carlo localization [68], appearance-based place recognition for topological localization [64], panoramic vision creation [64].

The problem of topological mobile localization has mainly three dimensions: a type of environment (indoor, outdoor, outdoor natural), a perception (sensing modality) and a localization model (probabilistic, basic). Numerous papers deal with indoor environments [17, 35, 63, 64] and a few deal with outdoor environments, natural or urban [24, 62].

Current work on place-based localization in indoor environments has focused on introducing probabilistic models to improve local feature matching and the integration of specific kernels. Different solutions to the third dimension of the problem, the localization model, have been developed over the past few years. These approaches first detect features and then compute a set of descriptors for these features. They either extract invariant features or they compute invariant descriptors based on non-invariant features. An obstacle to overcome becomes the finding of an adequate feature extraction solution.

Experimental results for wide baseline image matching suggest the need for local invariant descriptors of images. Invariant features are those that are invariant when exposed to a set of image transformations such as illumination, rotation, change of viewpoint. Earlier research into invariant features focused on invariance to rotation and translation. These methods have achieved relative success with object detection and image matching. Later work in invariant features has focused on expanding their invariance to illumination, scale and affine transforms. There has also been research into the development of fully invariant features [8, 43, 45]. In his milestone paper [38], D. Lowe has proposed a scale-invariant feature transform (SIFT) for recognition based on local extrema of difference-of-Gaussian filters in scale-space that is invariant to image scaling and rotation, illumination and viewpoint changes. Lately, a new method has been proposed, Affine-SIFT (ASIFT) that simulates all image views obtainable by varying the two camera axis orientation parameters, namely the latitude and the longitude angles, left over by the SIFT method [48]. However, full affine invariance has not been achieved due partly to the impractically large computational cost. SIFT is a 128 dimensional feature vector that captures the spatial structure and the local orientation distribution of a region surrounding a keypoint. Many studies have shown that SIFT is one of the best descriptors for keypoints [73].

A number of SIFT descriptor variants and extensions, including PCA-SIFT [30], GLOH (gradient location-orientation histogram) [44], SURF (speedup robust features) [3] have been developed ever since [20, 36, 71]. They claim more robustness and distinctiveness with scaled-down complexity, solving the issue of large computational costs. The SIFT method and its variants have been popularly applied for scene recognition [4, 15, 47] and detection [21, 50], robot localization [5, 49, 51], basic image matching problems [35].

The *Bag-of-Visual-Words* model was initially inspired by the bag-of-words models in text classification where a document is represented by an unsorted set of the contained words. This data modeling technique was first been introduced in the case of video retrieval [61]. Due to its efficiency and effectiveness, it became very popular in the fields of image retrieval and classification.

Classification of images rely either on unsupervised or supervised learning techniques. Categorizing in unsupervised learning scenarios is a much harder problem, due to the absence of class labels that would guide the search for relevant information. Problems of this kind have been rarely studied in the literature, mostly using probabilistic Latent Semantic Analysis (pLSA) [7] and Latent Dirichlet Allocation (LDA) [5, 60] to compute latent concepts in images from the co-occurrences of visual words in the collection. In supervised learning scenarios, image categorizing has been studied widely in the literature. Among supervised learning techniques, the most popular in this context are Bayesian classifiers [14, 23, 33, 34] and Support Vector Machines (SVM) [14, 33, 73]. [6] also uses random forests. Actually, state-of-the-art results are due to SVM classifiers: the method of [73], which combines a local matching of the features and specific kernels based on the Earth Movers Distance [55], yields the best results.

# Chapter 5

# Image Analysis

In this section, we describe the image features that have been used in this work in order to obtain a precise and effective model for the topological localization task. In order to obtain an image representation which captures the essential appearance of the location and is robust to occlusions and changes in image brightness we compare two different image descriptors and their associated distance measure. In the first case, we use image histograms integrated over the images and in the second case each image is represented by a set of local scale-invariant features.

Presented in detail in the famous D.H. Ballard and C.H. Brown book [2], the early processing techniques bring a collection of methods that exploit the imaging process at the basics. Filtering is a changing the image gray levels to enhance the visual appearance of images. These transformations increase the intensity between regions and are often dependent on object characteristics. For example, blurring the image at different levels, small intensity discontinuities will be erased and the accuracy of object retrieval will increase.

Template matching is a simple filtering technique which detects particular features in images. A similarity measure is computed in order to reflect how well the image matches the template.

Another important transformation of an image is exploring the gray levels (intensity of pixels) of an image by computing the frequencies of occurrence of each gray level in the image. An image histogram acts as a graphical representation of the tonal distribution in an image. It plots the number of pixels for each tonal value on the vertical axis with a particular brightness value on the horizontal axis. By analyzing the distribution of pixel amplitudes, you can gain some information about the visual appearance of an image. A high-contrast image contains a wide distribution of pixel counts covering the entire amplitude range. A low contrast image has most of the pixel amplitudes congregated in a relatively narrow range. A histogram accumulates the information and counts the number of image samples whose values lie within a given range of values, or bins. Usually, the wider histogram represents a more visually-appealing image.

A global histogram basically represents distribution of features. In a more general mathematical sense, a histogram is a function $h_i$ that counts the number of observations that fall into each of the disjoint categories (known as *bins*), whereas the graph of a histogram is merely one way to represent a histogram. Thus, if we let $n$ be the total number of observations and $k$ be the total number of bins, the histogram $h_i$ meets the following conditions:

$$n = \sum_{i=1}^{k} h_i \qquad (5.1)$$

A cumulative histogram is a mapping that counts the cumulative number of observations in all of the bins up to the specified bin. That is, the cumulative histogram $H_i$ of a histogram $h_j$ is defined as:

$$H_i = \sum_{j=1}^{i} h_j \qquad (5.2)$$

In conclusion, the histogram provides important information about the shape of a distribution. This basic processing technique will be detailed in the next subsection where our main focus are the features extracted form images. Also, the intensity of a pixel has the ability to bring more information. It is produced by the light reflected by a small area of surface near the corresponding point on the object. The perception of the object may be manipulated by the reflectivity properties. Surface orientation information can be obtained using intensity images as input and analyzing the constraints between neighboring surface elements. The neighboring orientations are relaxed against each other until each converges to a unique orientation. We exploited this technique in Section 5.2 where histograms of orientations brought us to a more robust model setting.

## 5.1 Global Features

Many recognition systems based on images use global features that describe the entire image, an overall view of the image that is transformed in histograms of frequencies. Adopting the analysis of global features has brought great improvement in robot localization systems as in [57] or in content based image retrieval systems as in medical related images analysis in [59]. Most shape and texture descriptors fall into this category. Such features are important because they produce very compact representations of images, where each image corresponds to a point in a high dimensional feature space.

One of the principal advantages of the global image coding lies in its computational efficiency: there is no need to parse the image or group its components in order to represent the spatial configuration of the scene. As a result, any standard classification technique can be used. On the other hand global features are sensitive to clutter and occlusion. As a result it is either assumed that an image only contains a single object, or that a good segmentation of the object from the background is available.

In our case, an image often does contain a scene and we are not interested in finding particular objects in the scene, but sometimes several organisms or particles are present. Towards our purpose, the common ground for using global features is to extract a signature for every image based on its pixel values, and to define a rule for comparing them. The signature can be shape, texture, color or any other information with which two images could be compared.

Global features capture the diagnostic structure of the image, an overall view of the image that is transformed in histograms of frequencies. Existing color-based general-purpose image retrieval systems as [10, 56] roughly fall into three categories depending on the signature extraction approach used: histogram, color layout, and region-based search. In this project, histogram-based search methods are investigated in two different color spaces. A color space is defined as a model for representing color in terms of intensity values. Typically, a color space defines a one-to-four-dimensional space. A color component, or a color channel, is one of the dimensions. Color spaces are related to each other by mathematical formulas.

However, color histograms have several inherent problems for the task of image indexing and retrieval. The first concern is their sensitivity to noisy interference such as lighting intensity changes and quantization errors. The second problem is their high dimensionality on representation. Even with coarse quantization over a chosen color space, color histogram feature spaces often occupy more than one hundred dimensions (i.e., histogram bins) which significantly increases the computation of distance measurement on the classification stage. Finally, color histograms do not include any spatial information and are therefore incompetent to support accurate image retrieval based on local image contents which will be needed in robot vizualization as a prerequisite.

In the following, we briefly describe several existing approaches that have been attempting to address these concerns. We attempt to model image densities using two different color spaces, RGB and HSV.

### 5.1.1 Color Space

A color space is defined as a model for representing color in terms of intensity values. A color component, or a color channel, is one of the dimensions. A color dimensional space (one dimension per pixel) represents the gray-scale space. Well-known color spaces include: RGB (**R**ed, **G**reen, and **B**lue, for display and printing processes), CMYK (**C**yan, **M**agenta, **Y**ellow, and **K**ey, in this case, meaning black, for television and video) YIQ (the color space used by the NTSC - **N**ational **T**elevision **S**ystem **C**ommittee - color TV system), YUV (standard set of primary colors typically used as part of a color image pipeline), HSL (**H**ue, **S**aturation, and **L**ightness), HSV (**H**ue, **S**aturation, and **V**alue). Although the number of existing color spaces is large, a number of these color models are correlated to intensity, linear combinations of RGB or normalized with respect to intensity rgb.

We concentrate on the following standard color features: RGB and HSV. In the sequel, we need to be precise on the definitions of RGB and HSV. Basically, in a preprocessing step, images are typically read as RGB models and then transformed into other color spaces.

### 5.1.2 RGB Color Model

The RGB (**R**ed, **G**reen, and **B**lue) color model is composed of the primary colors Red, Green, and Blue. They are considered the additive primaries since the colors are added together to produce the desired color. This system defines the color model that is used in most color CRT monitors and color raster graphics. To represent the RGB space, a cube can be defined on the R, G, and B axes. White is produced when all three primary colors at the maximum light intensity (255).

A color histogram denotes the joint probabilities of the intensities of the three color channels. The histogram is defined as:

$$h_{R,G,B}(r,g,b) = n \cdot Prob(R=r, G=g, B=b) \quad (5.3)$$

where R, G and B are the three color channels and $n$ is the number of pixels in the image. The color histogram is computed by discretizing the colors within the image and counting the number of pixels of each color. Since the number of colors is finite, it is usually more convenient to transform the three channel histogram into a single variable histogram. Given an RGB image, one transform is given by $t = r + n_r \cdot g + n_r \cdot n_g \cdot b$, where $n_r, n_g$ and $n_b$ are the number of bins for colors red, blue and green, respectively. This gives the single variable histogram:

$$h_T(t) = n \cdot Prob(T=t) \quad (5.4)$$

The RGB space has the major deficiency of not being perceptually uniform, this being the motivation of adding HSV color histograms.

### 5.1.3 HSV Color Model

This HSV (**H**ue, **S**aturation, and **V**alue) color model defines colors in terms of three constituent components: *hue*, *saturation* and *value* or *brightness*. The *hue* and *saturation* components are intimately related to the way human eye perceives color because they capture the whole spectrum of colors. The *value* represents intensity of a color, which is decoupled from the color information in the represented image. As *hue* varies from 0 to 1, the corresponding colors vary from red, through yellow, green, cyan, blue, and magenta, back to red, so that there are actually red values both at 0 and 1. As *saturation* varies from 0 to 1, the corresponding colors (hues) vary from unsaturated (shades of gray) to fully saturated (no white component). As *value*, or lightness/brightness, varies from 0 to 1, the corresponding colors become increasingly brighter.

This color model is attractive because color image processing performed independently on the color channels does not introduce false colors (hues). Furthermore, it is easier to compensate for artifacts and color distortions. For example, lighting and shading artifacts are typically isolated to the lightness channel. However, these color spaces are often inconvenient due to the necessary non-linearity in forward and reverse transformations with RGB space. We utilize a tractable form of the *hue*, *lightness* and *saturation* transform from RGB to HSV that has the above mentioned characteristics and is non-linear but easily invertible. The transformation from RGB to HSV is accomplished through the following equations: let the color triple $(r, g, b)$ define a color in RGB space and let $(h, s, v)$ be the transformed triple in HSV color space:

$$
\begin{aligned}
&r, g, b \in [0, 1] \\
\\
&h' = \left\{ \begin{array}{ll}
\frac{(g-b)}{\delta} & \text{if } r = \max(r, g, b) \\
\frac{2+(b-r)}{\delta} & \text{if } g = \max(r, g, b) \\
\frac{4+(r-g)}{\delta} & \text{if } b = \max(r, g, b)
\end{array} \right\} \\
\\
&h = h' \cdot 60 \\
&s = \frac{\max(r,g,b) - \max(r,g,b)}{\max(r,g,b)} \\
&v = \max(r, g, b) \\
\\
&h \in [0°, 360°] \\
&s, v \in [0, 1]
\end{aligned}
\tag{5.5}
$$

Finally, a quantization of the histograms is needed. Following Equation 5.1.2, we need to determine a proper histogram dimension (the number of histogram bins) $n$. This is already determined by the color representation scheme and quantization level. Most color spaces represent a color as a threedimensional vector with real values (e.g. RGB, HSV). We quantize the color space of three axes into $k$ bins for the first axis, $l$ bins for the second axis and $m$ bins for the third axis. The histogram can be represented as an $n$-dimensional vector where $n = k \cdot l \cdot m$. In general high resolution schemes, the histogram of RGB color space with $[255, 255, 255]$ range of three axes is represented as a 224-dimensional vector, HSV color space with $[360, 100, 100]$ range of three axes as a $3,600,000$ - dimensional vector. These high resolution representations, however, are unnecessarily large for image indexing. Because the retrieval performance is saturated when the number of bins is increased beyond some value, normalized color histogram difference can be a satisfactory measure of frame dissimilarity, even when colors are quantized into only 64 bins (4 Green × 4 Red × 4 Blue).

As a conclusion, we chosen a $18 \cdot 10 \cdot 10$ multidimensional HSV histogram, and a a $10 \cdot 10 \cdot 10$ multidimensional RGB histogram, as differences between colors of the office environment have a high level of similarity and have slight changes in hues.

## 5.2 Local Features

A different paradigm is to use local features, which are descriptors of local image neighborhoods computed at multiple interest points. The advantages of using these features are that they describe localized image regions (*patches*), the descriptors are computed around interest points, there is no need for segmentation and they are robust to occlusion and clutter. The disadvantage is that images are represented by different size sets of feature vectors and they do not lend themselves easily to standard classification techniques.

In this section, we describe typical ways in which local features are used. One of the key issues in dealing with local features is that there may be differing numbers of feature points in each image, making comparing images more complicated. Interest points are detected at multiple scales and are expected to be repeatable across different views of an object. The interest points are also expected to capture the essence of the object's appearance. The feature descriptor describes the image patch around an interest point.

The usual paradigm of using local features is to match them across images, which requires a distance metric for comparing feature descriptors. This distance metric is used to devise a heuristic procedure for determining when a pair of features is considered a match, e. g. by using a distance threshold. The matching procedure may also utilize other constraints, such as the geometric relationships among the interest points, if the object is known to be rigid. One advantage of using local features is that they may be used to recognize the object despite significant clutter and occlusion. They also do not require a segmentation of the object from the background, unlike many texture features, or representations of the object's boundary (shape features).

The main parts of localization processes explained in previous section are based on the analysis and matching of local image features. Choosing the feature is a very important practical issue, the purpose is to find the simplest and fastest feature that provides all the invariant properties required. There are many local features developed in the last years for image analysis, with the outstanding SIFT as the most popular. In the literature, there are several works studying the different features and their descriptors, for instance [36] evaluates the performance of the art in local descriptors, and [73] shows a study on the performance of different feature for object recognition.

We used different features along with the developed algorithms in our previous works [18, 19], to try to evaluate their efficiency for the aimed robotic tasks. The three kinds of features used in the experiments in next section are SIFT (Scale Invariant Feature Transform), ASIFT (Affine Scale Invariant Feature Transform) and RGB-SIFT (RGB Scale Invariant Feature Transform). Also, the localization experiments using these features show advantages and disadvantages of using one or another.

### 5.2.1 SIFT (Scale Invariant Feature Transforms)

The first descriptor that we consider is the scale-invariant (SIFT) features [8, 38, 39]. The SIFT features correspond to highly distinguishable image locations which can be detected efficiently and have been shown to be stable across wide variations of viewpoint and scale. The algorithm basically extracts features that are invariant to rotation, scaling an partially invariant to changes in illumination an affine transformations. The features generated should be highly distinctive. To aid the extraction of these features the SIFT algorithm applies a four stage filtering approach:

**Scale-Space Extrema Detection.** Interest points for SIFT features correspond to local extrema of Difference-of-Gaussian filters at different scales. The first step towards the detection of interest points is the convolution of the image with Gaussian filters at different scales which means generating progressively blurred out images, and the generation of Difference-of-Gaussian (DoG) images from the difference of adjacent blurred images. Convolution is an image filtering technique used for smoothing, such as Gaussian blurr or edge detection. The individual images are formed because of the increasing scale (the amount of blur). The number of octaves and scale depends on the size of the original image. However, the creator of SIFT suggests that 4 octaves and 5 blur levels are ideal for the algorithm.

Given a Gaussian blurred image:

$$L(x, y, \sigma) = G(x, y, \sigma) * I(x, y) \tag{5.6}$$

where

$$G(x, y, \sigma) = \frac{1}{2\pi\sigma} e^{-\frac{x^2+y^2}{2\sigma^2}} \tag{5.7}$$

is a variable scale Gaussian.

- $L$ is a blurred image
- $G$ is the Gaussian Blur operator
- $I$ is an image
- $x, y$ are the location coordinates
- $\sigma$ is the scale parameter (the amount of blur, D. Lowe [38] suggests $\sigma = 1.6$ close to optimal repeatability)
- The $*$ is the convolution operation in $x$ and $y$ (it *applies* gaussian blur $G$ onto the image $I$)

To efficiently detect stable keypoint locations in scale space, the Difference-of-Gaussian function convolved with the image, $DoG(x, y, \sigma)$ can be computed from the difference of two nearby scales separated by a constant multiplicative factor $k$:

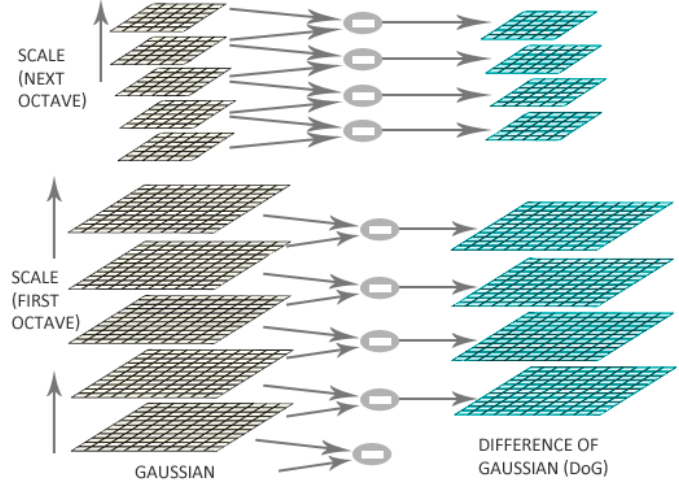

Figure 5.1: The process of computing the Difference-of-Gaussians. This figure has been taken from Lowe [38].

The result of convolving an image with a Difference-of-Gaussian filter:

$$\begin{aligned}\text{DoG}(x,y) &= (\text{G}(x,y,k\sigma) - \text{G}(x,y,\sigma)) * \text{I}(x,y) \\ &= \text{L}(x,y,k\sigma) - \text{L}(x,y,\sigma)\end{aligned} \quad (5.8)$$

Following this, the Difference-of-Gaussian function provides a close approximation to the scale-normalized Laplacian of Gaussian $\sigma^2 \nabla^2 G$ as studied by Lindeberg in [37] where he showed that the normalization of the Laplacian with the factor $\sigma^2$ is required for true scale invariance.

In detailed experiment settings, Mikolajczyk, in his Phd thesis [42] found that the maxima and minima of $\sigma^2 \nabla^2 G$ produce the most stable image features compared to a range of other possible image functions, such as the gradient, Hessian, or Harris corner function. The relationship between $DoG$ and $\sigma^2 \nabla^2 G$ can be understood from the heat equation (parameterized in terms of $\sigma$):

$$\frac{\partial G}{\partial \sigma} = \sigma^2 \nabla^2 G \quad (5.9)$$

and therefore

$$\text{G}(x,y,k\sigma) - \text{G}(x,y,\sigma) \approx (k-1)\sigma^2 \nabla^2 G \quad (5.10)$$

This shows that when the Difference-of-Gaussian function has scales differing by a constant factor it already incorporates the $\sigma^2$ scale normalization required for the scale-invariant Laplacian. The factor $k-1$ in the equation is a constant over all scales and therefore does not influence extrema location. The approximation error will go to zero as $k$ goes to 1, but in practice Lowe [38] has found that the approximation has almost no impact on the stability of extrema detection or localization for even significant differences in scale, such as $k = \sqrt{2}$. This computing of Difference-of-Gaussians locates edges and corners on the image. These edges and corners are necessary for finding keypoints.

The convolved images are grouped by octave (an octave corresponds to doubling the value of $\sigma$) as shown in Figure 5.1, and the value of $k$ is selected so that we obtain a fixed number of blurred images per octave. This also ensures that we obtain the same number of Difference-of-Gaussian images per octave. For each octave of scale space, the initial image is repeatedly convolved with Gaussians to produce the set of scale space images shown on the left. Adjacent Gaussian images are subtracted to produce the Difference-of-Gaussian images on the right. After each octave, the Gaussian image is down-sampled by a factor of 2, and the process repeated.

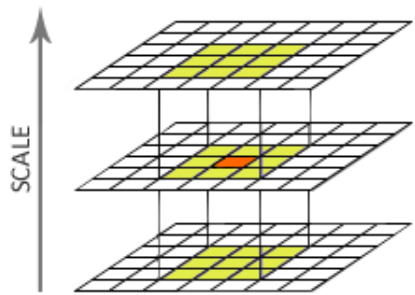

Figure 5.2: The detection of the maxima and minima of the Difference-of-Gaussian. This figure has been taken from Lowe [38].

**Keypoint Localization.** In order to detect the local maxima and minima of $DoG(x, y, \sigma)$, each sample point is compared to its eight neighbors in the current image and nine neighbors in the scale above and below as seen in Figure 5.2. The pixel marked is compared against its 26 neighbors in a $3 \times 3 \times 3$ neighborhood that spans adjacent DoGs Interest points (SIFT keypoints) are identified as local maxima or minima of the DoG images across scales. Each pixel in the DoGs is compared to its 8 neighbors at the same scale, plus the 9 corresponding neighbors at neighboring scales. If the pixel is a local maximum or minimum, it is selected as a candidate keypoint.

**Orientation Assignment.** For each candidate keypoint:

- Interpolation of nearby data is used to accurately determine its position
- Keypoints with low contrast are removed
- Responses along edges are eliminated

- The keypoint is assigned an orientation

To determine the keypoint orientation, a gradient orientation histogram is computed in the neighborhood of the keypoint (using the Gaussian image at the closest scale to the keypoint's scale). The contribution of each neighboring pixel is weighted by the gradient magnitude and a Gaussian window with a $\sigma = 1.6$ times the scale of the keypoint.

Peaks in the histogram correspond to dominant orientations. A separate keypoint is created for the direction corresponding to the histogram maximum, and any other direction within 80% of the maximum value. All the properties of the keypoint are measured relative to the keypoint orientation, this provides invariance to rotation. This approach contrasts with the orientation invariant descriptors is described in [58], in which each image property is based on a rotationally invariant measure.

**Keypoint Descriptor.** Once a keypoint orientation has been selected, the feature descriptor is computed as a set of orientation histograms on $4 \times 4$ pixel neighborhoods. The orientation histograms are relative to the keypoint orientation, the orientation data comes from the Gaussian image closest in scale to the keypoint's scale. Just like before, the contribution of each pixel is weighted by the gradient magnitude, and by a Gaussian with $\sigma = 1.6$.

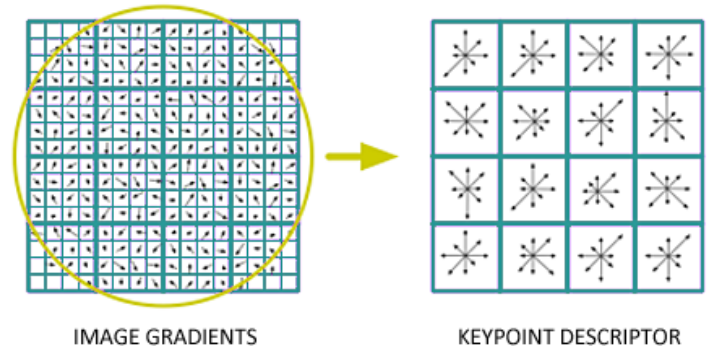

Figure 5.3: SIFT feature descriptor. This figure has been taken from Lowe [38].

The Figure 5.3 represents the initial image and the right one is filled with the extracted SIFT features. The features are visualized as circles with orientation lines showing the scale and orientation, and with oriented squares showing the sampling region. Histograms contain 8 bins each, and each descriptor contains an array of 4 histograms around the keypoint. This leads to a SIFT feature vector with $4 \times 4 \times 8 = 128$ elements. This vector is normalized to enhance invariance to changes in illumination.

A change in image contrast in which each pixel value is multiplied by a constant will multiply gradients by the same constant, so this contrast change will be canceled by vector normalization. A brightness change in which a constant is added to each image pixel will not affect the gradient values, as they are computed from pixel differences. Therefore, the descriptor is invariant to affine changes in illumination.

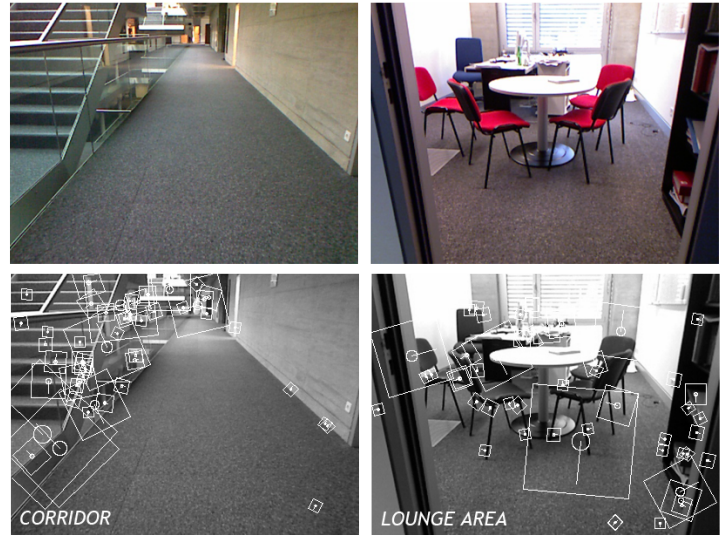

Figure 5.4: SIFT keypoints visualization

These resulting vectors are know as SIFT keys 5.4 and are used in a nearest-neighbors approach to identify possible objects in an image. Collections of keys that agree on a possible model are identified, when 3 or more keys agree on the model parameters this model is evident in the image with high probability. Due to the large number of SIFT keys in an image of an object, typically a $500 \times 500$ pixel image will generate in the region of 2000 features, substantial levels of occlusion are possible while the image is still recognized by this technique.

### 5.2.2 ASIFT (Affine Scale Invariant Feature Transforms)

The idea of combining simulation and normalization is the main ingredient of the SIFT method. The SIFT detector normalizes rotations and translations, and simulates all zooms out of an image. SIFT achieves the scale invariance by simulating the zoom in the scale-space. Because of this feature, it is the only fully scale invariant method.

As described in [48], ASIFT simulates with enough accuracy all distortions caused by a variation of the camera optical axis direction. Then it applies the SIFT method. In other words, ASIFT simulates three parameters: the scale, the camera longitude angle and the latitude angle and normalizes the other three (translation and rotation), what SIFT lacked.

To aid the extraction of these features we briefly describe another step before computing SIFT: Each image is transformed by simulating all possible affine distortions caused by the change of camera optical axis orientation from a frontal position. These distortions depend upon two parameters: the longitude $\varphi$ and the latitude $\theta$. The images undergo $\varphi$-rotations followed by tilts (latitude angle distortions) with parameter $t = \frac{1}{\cos\theta}$ (a tilt by t in the direction of x is the operation $u(x,y) \to u(tx,y)$).

These rotations and tilts are performed for a finite and small number of latitude and longitude angles, the sampling steps of these parameters ensuring that the simulated images keep close to any other possible view generated by other values of $\varphi$ and $\theta$.

The key observation is that, although a latitude angle distortion is irreversible due to its non-commutation with the blur, it can be compensated up to a scale change by digitally simulating a tilt of same amount in the orthogonal direction. As opposed to the normalization methods that suffer from this non-commutation, ASIFT simulates and thus achieves the full affine invariance. Against any prognosis, simulating the whole affine space is not prohibitive at all with the proposed affine space sampling. A two-resolution scheme will further reduce the ASIFT complexity to about twice that of SIFT.

### 5.2.3 RGB-SIFT (RGB Scale Invariant Feature Transforms)

RGB-SIFT descriptors are computed for every RGB channel independently. Therefore, each channel is normalized separately which brings another important aspect for SIFT, the invarince to light color change. For a color image, the SIFT descriptions independently from each RGB component and concatenated into a 384-dimensional local feature (RGB-SIFT) [9].

# Chapter 6

# Feature Matching

In Chapter 5 we introduced several different features to represent images in order to be able to decide whether two images are similar. In this chapter we introduce different dissimilarity measures to compare features. That is, a measure of dissimilarity between two features and thus between the underlying images is calculated. Many of the features presented are in fact histograms (color histograms, invariant feature histograms, texture histograms, local feature histograms, and other feature histograms). As comparison of distributions is a well known problem, a lot of comparison measures have been proposed and compared before [54].

Usually, local features from a pair of images are matched to produce a list of reliable point correspondences. The correspondences can then be used to perform image classification. In previous work by Lowe [8, 38, 39], image matching was performed by counting the number of vectors in the testing image that *matched* to vectors in the training image. Two vectors match if their Euclidean distance falls below a threshold. We decided to use the number of matches between two images as our similarity measure.

Distances used in computing histograms dissimilarities:

## 6.1 Bin-by-bin dissimilarity measures

Bin-by-bin comparison measures for histograms are usually thr primary solution for image mathing because they can be computed very fast. A drawback is that due to the fact that they only compare bin-by-bin they cannot be accounted for similarities between the underlying values of the bins and this can became a problem for an entire system. For example, in Figure 6.1 the histograms a) and b) have the same distance as the histograms a) and c), but obviously a) and b) should be more similar if they were gray value histograms.

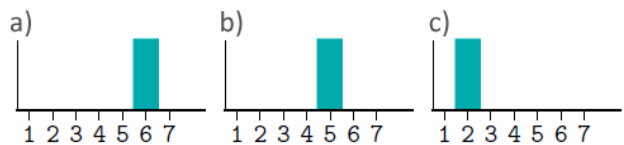

Figure 6.1: Bin-by-bin similarity distance computation sample

In the following, dissimilarity measures to compare two histograms $H$ and $K$ are proposed. Each of these histograms has $n$ bins and $H_i$ is the value of the $i$-th bin of histogram $H$.

- **Minkowski-form Distance** ($L_1$ distance is often used for computing dissimilarity between color images, also experimented in color histograms comparison [32]):

$$\mathrm{D}_{Lr}(H, K) = (\sum_{i=1} |H_i - K_i|)^{\frac{1}{r}} \qquad (6.1)$$

- **Kullback-Leibler Divergence** (from the information theory point of view, the K-L divergence has the property that it measures how inefficient on average it would be to code one histogram using the histogram):

$$\mathrm{D}_{KL}(H, K) = \sum_{i=1} \log \frac{H_i}{K_i} \qquad (6.2)$$

- **Jensen Shannon Divergence** (also referred to as **Jeffrey Divergence** [16], is an empirical extension of the Kullback-Leibler Divergence. It is symmetric and numerically more stable):

$$\mathrm{D}_{JSD}(H, K) = \sum_{i=1} H_i \log \frac{2H_i}{H_i + K_i} + K_i \log \frac{2K_i}{K_i + H_i} \qquad (6.3)$$

- **$\chi^2$ Distance** (measures how unlikely it is that one distribution was drawn from the population represented by the other, [46]):

$$\mathrm{D}_{\chi^2}(H, K) = \sum_{i=1} \frac{(H_i - K_i)^2}{H_i} \qquad (6.4)$$

- **Bhattacharyya Distance** [11] (measures the similarity of two discrete or continuous probability distributions). For discrete probability distributions H and K over the same domain, it is defined as:

$$\mathrm{D}_B(H, K) = -\ln \sum_{i=1} \sqrt{(H_i K_i)} \qquad (6.5)$$

## 6.2 Cross-bin dissimilarity measures

Since the distance measures described so far neglect similarities between different bins of the histograms, even small changes in color or lighting conditions may lead to major changes in histogram distances. The measures described in the remainder of this chapter are developed to overcome this problem. Similarities between the underlying values represented by different bins are taken into consideration.

- **Earth Movers Distance (EMD)** [55] (reflects the minimal amount of work that has to be performed to transform one distribution into the other by shifting portions of the distribution between bins. This is a special case of the transportation problem):

$$\mathrm{D}_{EMD}(H, K) = \frac{\sum_{i,j} h_{i,j} k_{i,j}}{\sum_{i,j}} \qquad (6.6)$$

  Here $h_{i,j}$ denotes the dissimilarity between bin $i$ and bin $j$ and $k_{i,j}$ 0 is the optimal flow between the two distributions such that the total cost $c_{i,j} = h_{i,j} k_{i,j}$ is minimized. The following constraints have to be taken into account for all $i, j$:

$$\begin{array}{l} \sum_i k_{i,j} \leq K_j \\ \sum_j k_{i,j} \leq H_i \\ \sum_{i,j} k_{i,j} = \mathrm{D}(H_i, K_j) \end{array} \qquad (6.7)$$

  A major advantage of the EMD is that each image may be represented by a histogram with individual binning.

- **Match distance** (the match distance between two one-dimensional histograms defined as the $L1$ distance between their corresponding cumulative histograms. For one-dimensional histograms with equal areas, this distance is a special case of EMD with the important difference that the match distance cannot handle partial matches):

$$\begin{array}{l} \mathrm{D}_M(H, K) = \sum_{i=1} |h_i - k_i| \\ \mathrm{h_i} = \sum_{j \leq i} h_j \end{array} \qquad (6.8)$$

# Chapter 7

# Bag-of-Visual-Words (BoVW)

Recent advances in the image recognition field have shown that **bag-of-visual-words** [14, 23] - a strategy that draws inspiration from the text retrieval community - approaches are a good method for many image classification problems. **BoVWs** representations have recently become popular for content based image classification because of their simplicity and extremely good performance.

Being inspired from natural language processing field, the documents of text contain distribution of words, and thus can be compactly summarized by their word counts (known as a *bag-of-words*). Since the occurrence of a given word tends to be sparse across different documents, an index from a dictionnary that maps words to the files in which they occur can take a keyword query and immediately produce relevant content. Being a categorization technique inspired, it includes normalization and feature selection for generating image representations with different dimensions of visual words and study their effectiveness in image classification tasks (more exactly on topological image information which implies interior scenes).

Linking to computer vision, this data modeling technique was first been introduced in the case of video retrieval [61]. Recently, this technique had also great success in medical image retrieval [25] or in different approaches and experiments in [1, 72]. Due to its efficiency and effectiveness, it became very popular in the fields of image retrieval and classification.

Basically, in computer vision, an image is represented by an unsorted set of discrete **visual words**, which are obtained from different descriptors. Therefore, a large corpus of representative images will populate the feature space with descriptor instances. Next, the sampled features are clustered in order to quantize the space into a discrete number of **visual words**. The **visual words** are the cluster centers, denoted with the large green circles. Now, given a new image, the nearest **visual word** is identified for each of its features. This maps the image from a set of high-dimensional descriptor to a list of **word** numbers. A **bag-of-visual-words** histogram is used to summarize the entire image. It counts how many times each of the visual words occurs in the image.

Therefore, the vocabulary building algorithm contains three components: the *patch* identifier, the visual appearance descriptor, and the quantization algorithm. The first two stepts were already explained in Chapter 5.

The vocabulary enables a compact reduction of all image's interest points.

The basic idea is to treat images as loose collections of independent patches, sampling a representative set of patches from the image, evaluating a visual descriptor vector for each patch independently, and to use the resulting distribution of samples in descriptor space as a characterization of the image. The image's empirical distribution of words is captured with a histogram counting how many times each word in the visual vocabulary occurs within it.

Local image descriptors are high-dimensional, real-valued feature points. Their discretisation is an important step with resulting data that will be used in further stages. To do so, we must impose a quantization on the feature space of local image descriptors. That way, any novel descriptor vector can be coded in terms of the (discretized) region of feature space to which it belongs. The standard pipeline to form a so-called *visual vocabulary* consists of collecting a large sample of features from a representative corpus of images, and quantizing the feature space according to their statistics.

More exactly, in the training phase, local feature descriptors are found for each training image, as described above. The vocabulary construction relies on an incremental nearest neighbor classifier. For a new feature vector, we perform a $L_1$-normalization and find the closest *visual word* in the current vocabulary. If the distance between the word and the feature is below a threshold, the *visual word* is recognized, otherwise a new word initialized to the feature position is added. This method is clearly sensitive to noise in the feature extraction and to the order of feature processing, problems that are solved when using a batch method such as $k$-means. For each training image descriptor, the nearest *visual word* are found, and a histogram of how many times a *visual word* is found in the vocabulary is produced and it is also $L_1$-normalized.

We consider the *BoVW* representation for images, with each image being represented by a collection of local descriptors. We denote by $N$ the number of training images, and by $X_i = (x_i^1, \ldots x_i^n)$ the collection of keypoints used to represent image $I_i$ where $x_i^j \in \chi, j = 1, \ldots n_i$, $n$ the length of the descriptor is a keypoint in feature space $\chi$.

Basically, to give an estimation of the distribution we create histograms of the local features. To circumvent this problem we apply a two step processing: first, a clustering algorithm is applied to a reasonably large set of local features. The obtained partitioning allows us to represent all local features by a cluster number, thus discretizing the local features by using a vocabulary.

To create a local feature histogram for an image, the local features are extracted and for each of the local features the cluster representing it best is determined. Then a histogram of these cluster memberships is created. This process allows adjusting the amount of data easily. By creating many cluster centers a large histogram is yielded and by creating only a few cluster centers smaller histograms are yielded. The histogram has the same number of bins as the partitioning has clusters. A view of this process is depicted in Figure 7.1.

The key idea of the *bag-of-visual-words* representation is to quantize each keypoint into one of the *visual words* that are often derived by clustering. Typically $k$-means clustering is used. The size of the vocabulary $k$ is a user-supplied parameter. The *visual words* are the $k$ cluster centers. The baseline of our tests are based on a bag-of-visual-words with a 100 *visual words*, meaning a 100-means clustering. The resulting $k$ $n$-dimensional cluster centers $c_j$ represent the *visual words*.

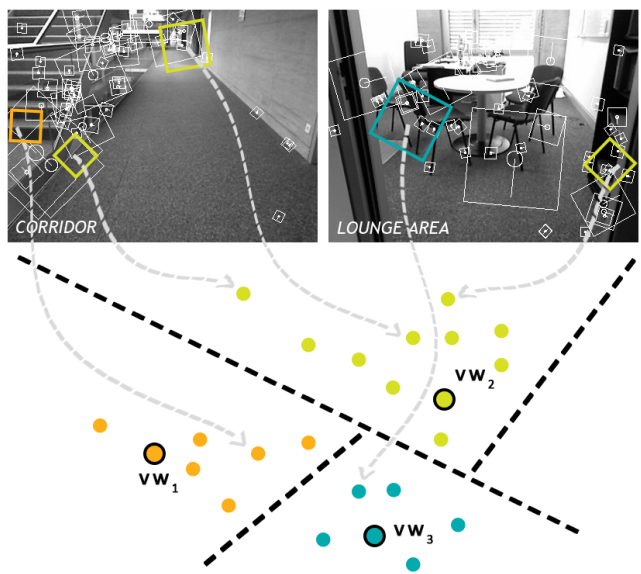

Figure 7.1: Visual Vocabulary

We use a vector quantization technique which clusters the keypoint descriptors in their feature space into different numbers of clusters using a diverse distance-based $k$-means clustering algorithm and encodes each keypoint by the index of the cluster to which it belongs. The obtained vectors have the size of the number of clusters that will compose the vocabulary. Choosing the right vocabulary size involves a trade-off discrimination and generalization. Once we have the vocabulary build, every keypoint is mapped on the feature space.

To obtain the histogram $h_i$ for image $I_i$ we perform vector quantization on all its descriptors $x_i^j$ onto the cluster centers $c_j$: Each $x_i^j$ is assigned the label $j$ of the closest cluster, using different distance measures (Chapter ..) in the feature space $\chi$. Using histograms of local features is motivated by the fact that for image retrieval good response times are required and this is hard to achieve using the huge amount of data incorporated in approaches using local feature queries. Here the amount of data is reduced by estimating the distribution of local features for each of the images.

Forming a histogram over all the labels encountered within an image yields the $k$-bin histogram $h_i$, which is normalized to sum to 1. Samples of images histograms, applied on our dataset, are represented in Figure 7.2. Let $h_i$ denote the normalized number of keypoints in image $I_i$ that are mapped to visual word $c_l$. Given $m$ visual words, $h_i$ is computed as:

$$h_i = \frac{1}{n_i} \sum_{j=1}^{n_i} f_k(x_i^j) \qquad (7.1)$$

$$f_k(x_i^j) = \left\{ \begin{array}{ll} 1 & , \|x_j - c_i\| \leq \|x_j - c_l\|, l = 1, \ldots |V| \text{ and } i \neq l \\ 0 & , \text{otherwise} \end{array} \right\} \qquad (7.2)$$

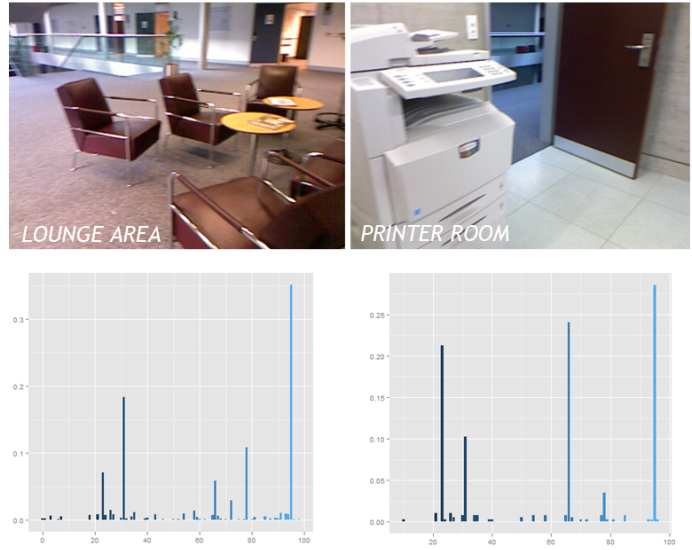

Figure 7.2: Sample Histograms of SIFT features mapped on our visual words

$h_i$ is the number of descriptors in image $I_i$ having the closest distance to the $i$-th visual word $v_i$ and $k$ is the total number of descriptors in image $I_i$. $f_{f_j}^i$ is equal to one if the $j$-th descriptor $x_j$ in image $I_i$ is closest to visual word $c_i$ among other visual words in the vocabulary $V$. The bag-of-words representation for image $I_i$ is expressed by vector $h_i = (h_i^1, \ldots h_i^n)$

Once the vocabulary is established, the corpus of sampled features can be discarded. Then novel images features can be translated into words by determining which *visual word* they are nearest to in the feature space. The interest points and corresponding descriptors are extracted for the query image $I_j$ and mapped to the visual words. The histogram $h_j$ based on the cluster mapping is used in a nearest neighbor query using the distance against the database of previous stored histograms.

# Chapter 8

# Experiments

In this chapter, we explain the experimental setup, then we present and discuss the results. The different choices of distance measures and classification parameters are analyzed performing also a comparison with previous work results. Conclusions are drawn in benefit of an accurate solution for topological localization, data modeling and classification.

## 8.1 Datasets (Benchmark)

The chosen dataset contains images from nine sections of an office obtained from **CLEF (Conference on Multilingual and Multimodal Information Access Evaluation)**. Detailed information about the dataset are in the overview and ImageCLEF publications [40, 53, 57]. This dataset contains images that are widely used in topological localization image classification papers and a preview is presented in Figure 8.1.

The dataset has already been split into three training sets of images, as shown in Table 8.1 one different from another, as seen in Table 8.1. The provided images are in the RGB color space. The sequences are acquired within the same building and floor but there can be variations in the lighting conditions (sunny, cloudy, night) or the acquisition procedure (clockwise and counter clockwise).

| Areas | # Images | | |
|---|---|---|---|
| | training1 | training2 | training3 |
| Corridor | 438 | 498 | 444 |
| ElevatorArea | 140 | 152 | 84 |
| LoungeArea | 421 | 452 | 376 |
| PrinterRoom | 119 | 80 | 65 |
| ProfessorOffice | 408 | 336 | 247 |
| StudentOffice | 664 | 599 | 388 |
| TechnicalRoom | 153 | 96 | 118 |
| Toilet | 198 | 240 | 131 |
| VisioConference | 126 | 79 | 60 |

Table 8.1: Training Sequences of An Office Environment

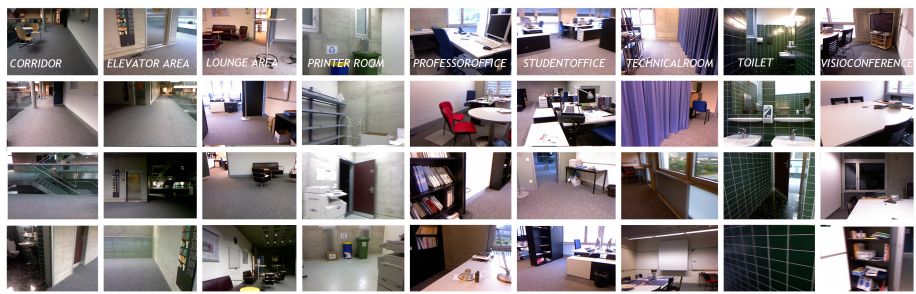

Figure 8.1: Dataset Preview

## 8.2 Classification

Many of the features presented in Chapter 5 are in fact histograms (color histograms, invariant feature histograms, texture histograms, local feature histograms). As comparison of distributions is a well known problem, a lot of comparison measures have been proposed in Chapter 6 and compared. To analyze the different measure distances we summarize a well known choice for supervised classification.

**Support Vector Machines** were first introduced by Vapnik and Chervonenkis in [66]. The core idea is to find the optimal hyperplane to separate a dataset, while there are theoretically infinite hyperplanes to separate the dataset. A hyperplane is chosen, so that the distance to the nearest point of both classes is maximized. The points spanning the hyperplane are the *support vectors*, hence the name Support Vector Machines [13].

SVMs are state-of-the-art large margin classifiers which have recently gained popularity within visual pattern and object recognition [14, 30, 33, 67, 69, 73]. In this section we provide a brief review of the theory behind this type of algorithm [41, 65].

Consider the problem of separating the set of training data $(x_1, y_1), (x_2, y_2) \ldots (x_m, y_m)$ into two classes, where $x_i \in \Re$ is a feature vector and $y_i \in (-1, +1)$ its class label. If we assume that the two classes can be separated by a hyperplane $w \cdot x + b = 0$ in some space $H$, and that we have no prior knowledge about the data distribution, then the optimal hyperplane is the one which maximizes the margin [65]. The optimal values for $w$ and $b$ can be found by solving a constrained minimization problem, using Lagrange multipliers $\alpha_i (i = 1, \ldots, m)$. The ($x_i$ are the *support vectors*. $K(x, y) = x \cdot y$ (kernel function) corresponds to constructing an optimal separating hyperplane int he input space $\Re_N$.

$$f(x) = sgn \left( \sum_i^m \alpha_i y i K(x_i, x) + b \right) \tag{8.1}$$

Choosing the most appropriate kernel highly depends on the problem at hand - and fine tuning its parameters can easily become a tedious task. Automatic kernel selection is possible and is discussed in [27]. The choice of a kernel depends on the problem because it depends on what data has to be modeled. A *polynomial kernel*, for example, allows to model feature conjunctions up to the order of the polynomial. *Radial basis* function allows to pick out circles (or hyperspheres) - in constrast with the *linear kernel*, which allows only to pick out lines (or hyperplanes). The motivation behind the choice of a particular kernel can be very intuitive and straightforward depending on what kind of information we are expecting to extract about the data [41].

**The Linear Kernel** is the simplest kernel function. It is given by the inner product $\langle x, y \rangle$ plus an optional constant $c$. Kernel algorithms using a linear kernel are often equivalent to their non-kernel counterparts.

$$K_l(x, y) = x^T y + c \tag{8.2}$$

**The Gaussian Kernel** is an example of radial basis function kernel.

$$K_g(x, y) = \exp\left(-\frac{\|x - y\|^2}{2\sigma^2}\right) \tag{8.3}$$

The adjustable parameter $\sigma$ plays a major role in the performance of the kernel, and should be carefully tuned to the problem at hand. If overestimated, the exponential will behave almost linearly and the higher-dimensional projection will start to lose its non-linear power. In the other hand, if underestimated, the function will lack regularization and the decision boundary will be highly sensitive to noise in training data.

**The $\chi^2$ Kernel** comes from the $\chi^2$ distribution.

$$K_{\chi^2}(x, y) = 1 - \sum_{i=1}^{n} \frac{(x_i - y_i)^2}{\frac{1}{2}(x_i + y_i)} \tag{8.4}$$

The kernel function provides a cheap way to equivalently transform the original point to a high dimension and perform the quadratic optimization in that high dimension space.

SVM is a highly effective model and works well across a wide range of problem sets. Although it is a binary classifier, it can be easily extended to multi-class classification by training a group of binary classifiers and using one-vs-all or one-vs-one to predict. SVM is a very powerful technique and perform good in a wide range of non-linear classification problems. It works best when there is a small set of input features because it will expand the features into higher dimension space, providing that there is a good size of training data (otherwise, overfitting can occur) which can lead to a feature selection process.

## 8.3 Vocabulary Size Evaluation

In order to apply a *bag-of-visual-words* method, the size of the vocabulary $k$ must be supplied as an important parameter. Further perfomance evaluation tests depend on this establishment. Since we aim to an effective system, the vocabulary size needs to meet the requirement of a clustering process to have the smallest computing time as it can get. Table 8.2 shows the different visual vocabulary sizes and their computing time.

As we mentioned before in Chapter 7, the *visual words* are the $k$ cluster centers. The quantization technique which clusters the keypoint descriptors in their feature space into different numbers of clusters uses a distance-based $k$-means clustering algorithm and encodes each keypoint by the index of the cluster to which it belongs. The obtained vectors have the size of the number of clusters that will compose the vocabulary. Choosing the right vocabulary size involves a trade-off discrimination and generalization and also becomes an important center of attention of many previous works [14, 28, 29, 70].

| Method | Time |
|---|---|
| 100 | 134 minutes 23 seconds |
| 200 | 216 minutes 57 seconds |
| 500 | 345 minutes 17 seconds |
| 1000 | 520 minutes 45 seconds |
| 2000 | 1234 minutes 14 seconds |

Table 8.2: Time Comparison for Vocabulary Generation

The baseline of our tests are based on a *bag-of-visual-words* with 100 *visual words*, meaning a 100-means clustering. The resulting 100 $n$-dimensional cluster centers $c_j$ represent the *visual words*.

## 8.4 Performance Evaluation for Histogram Comparison

The objective of this section is twofold: on one hand, we want to validate different configurations and proposed methods for indoor localization and, on the other hand, we target to experimentally determine if using different features, metrics and classifiers improves significantly the localization results, establishing from these results new proposals for image recognition.

Although successive images acquired by the robot, the temporal and route continuity, could be used to incrementally refine the localization, in our experiments, we evaluated a more challenging task, the task of detecting the location without using this information. In the case of a *kidnapping* situation or having only a single image available for localization, the robot could perform more efficient in recognizing the environment. The robustness of the system config-urations are evaluated and conclusions are drawn in the advantage of a more accurate recognition.

Common evaluation measures are used:

- **Recall.** A measure of the ability of a system to present all relevant items.

$$R = \frac{number\ of\ relevant\ items\ retrieved}{number\ of\ relevant\ items\ in\ collection} \tag{8.5}$$

- **Precision.** A measure of the ability of a system to present only relevant items.

$$P = \frac{number\ of\ relevant\ items\ retrieved}{total\ number\ of\ items\ retrieved} \tag{8.6}$$

- **F-measure.** A measure of a system accuracy. It considers both the precision $P$ and the recall $R$ of the test to compute the score.

$$F = 2\frac{P \cdot R}{P + R} \tag{8.7}$$

- **Error Rate.** A measure of the ability of a system to present irrelevant items.

$$ER = \frac{no.\ of\ relevant\ items\ in\ collection - no.\ of\ relevant\ items\ retrieved}{number\ of\ relevant\ items\ in\ collection} \tag{8.8}$$

To evaluate the features choices and measure distances, precision was plotted against recall in Precision/Recall (P/R) graphs and also, similarity values were represented in confusion matrices.

### 8.4.1 Comparison of Different Distance Functions for Global Features

RGB and HSV color histograms are subject to tests with Jeffrey Divergence 6.1, $\chi^2$ 6.1, Bhattacharyya 6.1, Minkowski 6.1 and respectively the widely used Euclidean distance measures. These were chosen considering the literature that underlies them as achieving the best results in image matching [11, 54].

The retrieved classes for images (*Corridor*, *LoungeArea* etc.) depend on a threshold, those below this value being rejected. This becomes an optimization problem of finding the best value that will cut the unwanted results, considering that it is better to have no results than inconsistent results. To accomplish this, we used the genetic algorithm explained in detail in [22]. For these experiments, we used a population of 200 individuals, the mutation probability of 0.15, and the crossover, of 0.7. The optimization process is stopped after 1000 generations. We used a selection scheme *rank selection* with *elitism*.

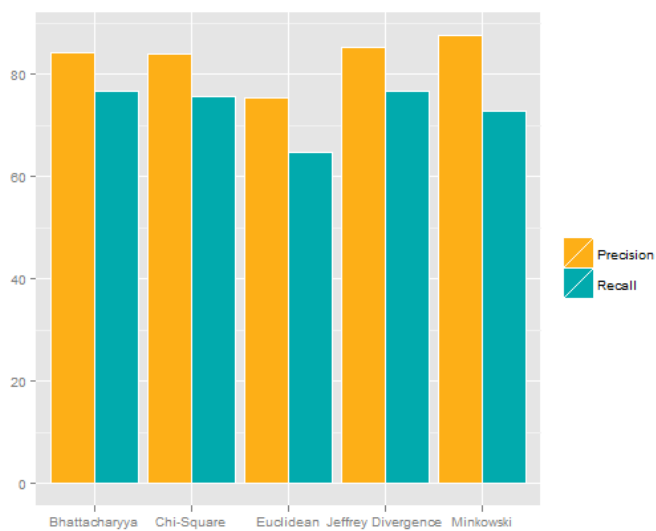

Figure 8.2: Precision and recall depending on measure distance (RGB Histograms)

For RGB histograms, as can be seen in Figure 8.2, Bhattacharyya 6.1 and Jeffrey Divergence 6.1 obtained the highest recall and also, high precision, the highest F-measure being obtained by Jeffrey Divergence (0.806) extremely close to Bhattacharyya (0.802). The lowest performance is with Euclidean distance, having not only a low recall which means that this solution will bring more irrelevant results than using the other distances, but also a lower precision.

In the case of using HSV histograms, Figure 8.3, the Bhattacharyya distance leaded to good results with a F-measure of 0.81 close to $\chi^2$ distance with 0.807

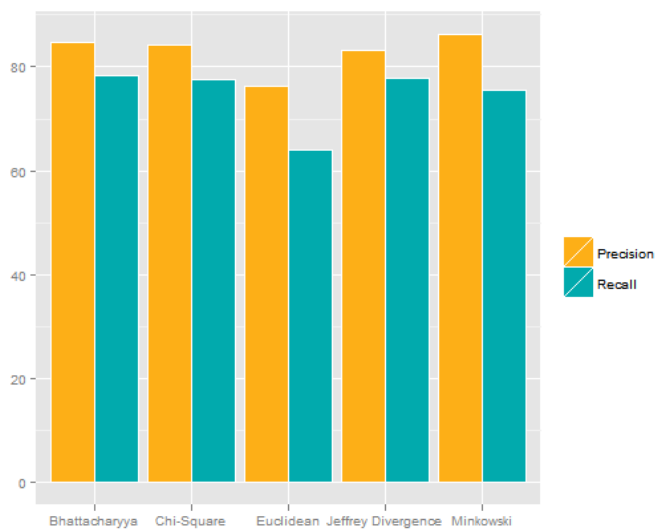

Figure 8.3: Precision and recall depending on measure distance (HSV Histograms)

and Minkowski 6.1 with 0.805 F-measure.

Folowing these chosen metrics, we adopted the vizualization with confusion matrices. A confusion matrix is an array showing relationships between true and predicted classes. Entries on the diagonal of the matrix, in blue, count the correct calls. Entries off the diagonal, in fading blue, count the misclassifications.

Corresponding to the confusion matrix represented in Figure 8.6 and Figure 8.7, the results show that HSV histogram with Bhattacharyya 6.1 distance yielded very similar results with RGB choices of distances (Figure 8.4 and Figure 8.5) but clearly outperforms RGB histogram comparison with Jeffrey Divergence 6.1 distance, similarity probabilty peaking at 100% in some of the office sections (*PrinterRoom, StudentOffice*).

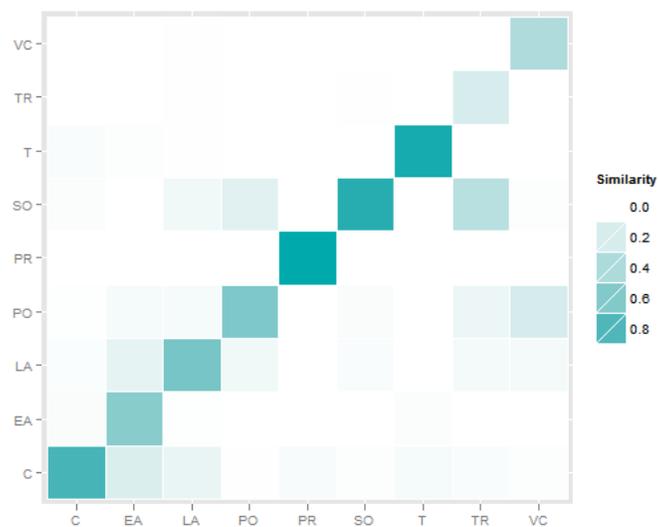

Figure 8.4: Confusion Matrix (RGB Histograms) using Bhattacharyya Distance

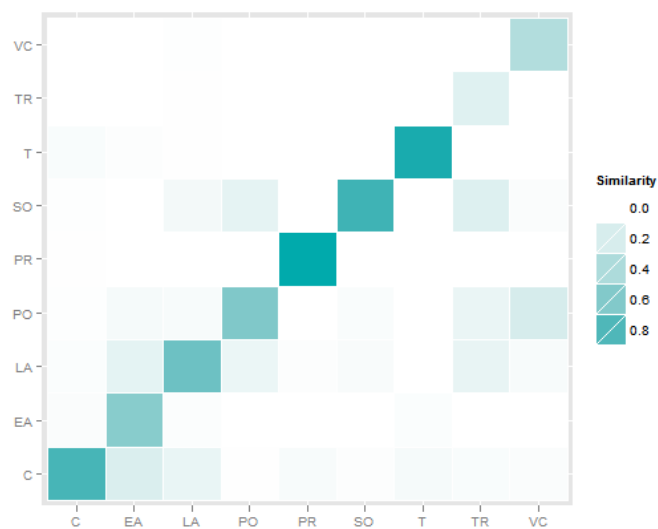

Figure 8.5: Confusion Matrix (RGB Histograms) using Jeffrey Divergence Distance

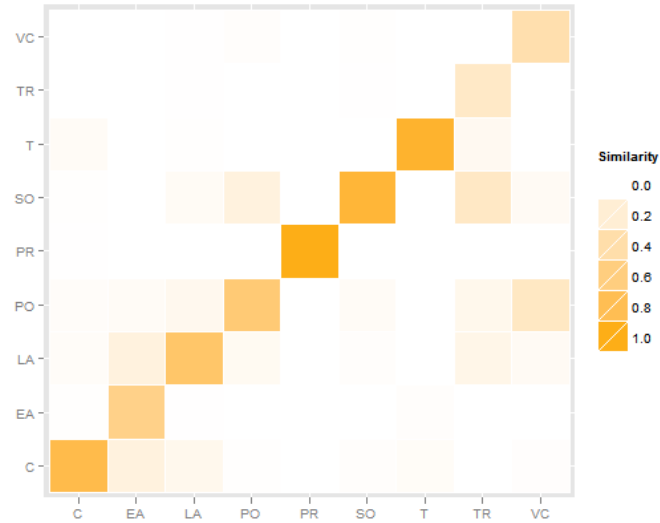

Figure 8.6: Confusion Matrix (HSV Histograms) using Bhattacharyya Distance

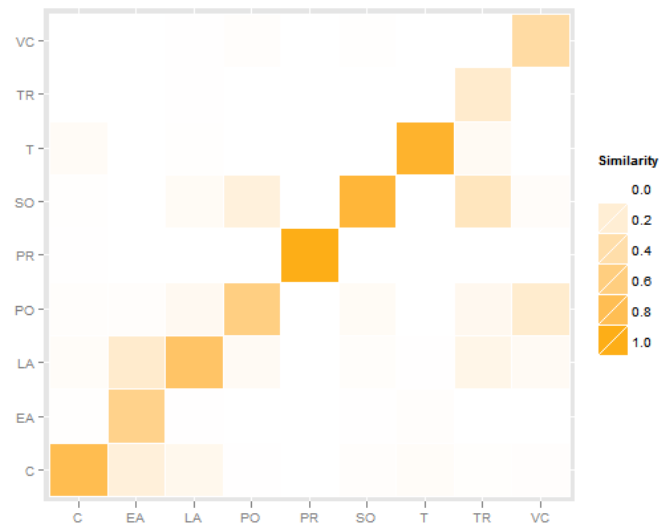

Figure 8.7: Confusion Matrix (HSV Histograms) using $\chi^2$ Distance

### 8.4.2 Comparison of Different Distance Functions for Local Features

These results were obtained performing experiments on local feature histograms obtained using the *bag-of-visual-words* model. The descriptors are quantized and normalized. Different dissimilarity measures for the different types of features are compared experimentally and the performance for the different types of features is quantitatively measured.

Even though the various dissimilarity measures described in Chapter 6 are applicable, we chose literature-based known as having the best results: Euclidean, Minkowski 6.1, $\chi^2$ and Jeffrey Divergence 6.1 distances. For each of the local features descriptors presented in Section 8.4.2, we created Precision/Recall graphs from which we determine the superior runs. Figures 8.8, 8.10, 8.9 show the Precision/Recall graphs for SIFT, ASIFT and RGB-SIFT apart.

These graphs provide an immediate, visual sense of the comparison of local quantized features' overall performance. Curves closest to the upper right-hand corner of the graph (where recall and precision are maximized) indicate the best performance. Comparisons are best made in three different recall ranges: 0 to 0.2, 0.2 to 0.8, and 0.8 to 1. These ranges characterize high precision, middle recall, and high recall performance, respectively. The Precision/Recall curves in Figures 8.8, 8.10, 8.9 show that there is still vast room for improvement but the most promising results were obtained in the case of the usage of SIFT descriptors with Minkowski 6.1 and Euclidean distance.

The results show that Euclidean and Minkowski 6.1 distance yielded very similar results, in the case of SIFT features matching.

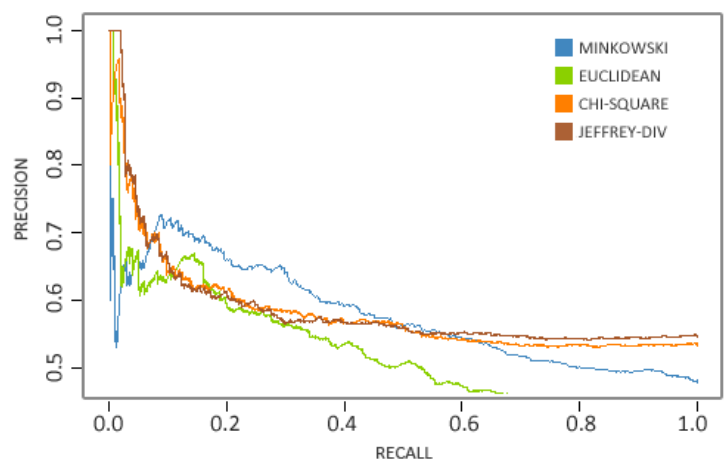

Figure 8.8: PR curves using different distance measures (ASIFT)

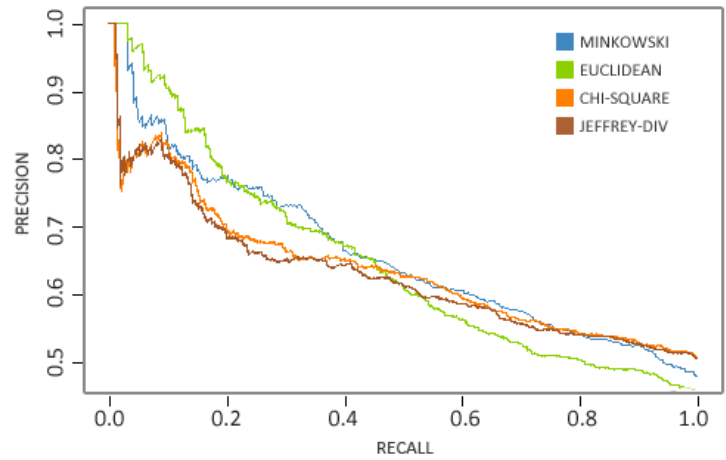

Figure 8.9: PR curves using different distance measures (RGB-SIFT)

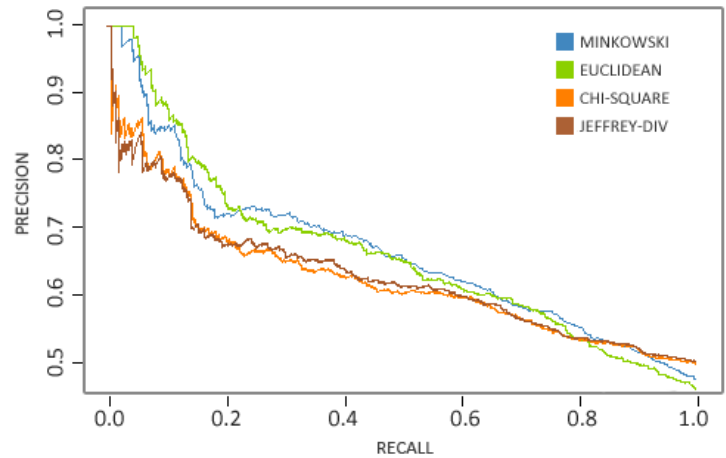

Figure 8.10: PR curves using different distance measures (SIFT)

## 8.5 Performance Comparison for Topological Localization

Since we investigated different types of features and distance functions we are interested in the discrimination performances of these comparing performance to see which features/dissimilarity measures lead to good results and which do not.

First, different distance measures for the different types of features were compared in Sections 8.4.2 and 8.4.1.

As it is well known that combinations of different methods lead to good results [31], an objective is to combine the presented features. However, it is not obvious how to combine which features. To analyze the characteristics of features and which features have similar properties, we perform an evaluation on selected configurations as shown in Table 8.3.

Table 8.3 presents results for different training and test situation. Training data was already split into three different sets of images as mentioned in Table 8.1. The evaluation was perfomed choosing *Training 1* and *3* (Table 8.1) for training and *Training 2* for testing.

| Method | R[%] | P[%] | ER[%] | F |
|---|---|---|---|---|
| RGB-Only | 73.73 | 82.02 | 10.11 | 0.77 |
| HSV-Only | 76.46 | 82.34 | 7.14 | 0.79 |
| RGB-HSV | 76.42 | 79.66 | 4.06 | 0.780 |
| Basic-BoVW-SIFT | 45.10 | 46.51 | 3.04 | 0.45 |
| Basic-BoVW-SIFT+HSV+RGB | 76.85 | 79.26 | 3.06 | 0.780 |
| Basic-BoVW-ASIFT+HSV+RGB | 77.60 | 79.97 | 2.96 | 0.787 |
| SVM-RBF-BoVW-SIFT+HSV+RGB | 78.87 | 78.87 | 0.0 | 0.788 |
| SVM-LINEAR-BoVW-SIFT+HSV+RGB | 78.63 | 78.94 | 0.39 | 0.787 |
| SVM-$\chi^2$-BoVW-SIFT+HSV+RGB | 78.43 | 78.52 | 0.11 | 0.784 |

Table 8.3: Performance Comparison for Topological Localization

The first column gives a description of the used training method. The descriptions of the configurations are straight forward, for example, *Basic-BoVW-SIFT+HSV+RGB* means a configuration of a combination of RGB and HSV color histograms and *Basic-BoVW-SIFT* a bag of visual words formed with SIFT feature vectors. The chosen measure distances were decided in Section 8.4.1, Jeffrey Divergence 6.1 for RGB histograms, Bhattacharyya 6.1 for HSV histograms and Minkowski for SIFT feature vectors. The second column gives the recall values for the training data, the third - the precisions and the forth gives the error rates. The F-measure is computed and representated in the fifth column of the table. The table also shows that feature selection only is not sufficient to increase the recognition rate but more flexibility is needed here, this fact led to different combinations.

The $RGB - Only$ gives an absolute baseline error rate, also $HSV - Only$

a very close error rate. The error rate decreases in the combination of the simple features $RGB - HSV$ and it is already sufficient to obtain very good results. In the case of local features, the fact that SIFT feature matching with Minkowski 6.1 and Euclidean outperforms ASIFT feature matching with Minkowski distance was observed earlier in this work in Section 8.4.2.

The results improve with the addition of the SVM classification step and We emphasize the fact the *SVM-RBF-BoVW-SIFT+HSV+RGB* has an error rate of 0%, meaning that all the retrieved images are also relevant. We also add the observation that a SVM classification of SIFT mapped on visual words can get to a maximum of 50% accuracy, but these results are very assuring in the context of a configuration in which are implied the usage of other feature descriptors. Thereby, the configuration that combines SIFT words, HSV and RGB histograms and a classification with a SVM with RBF kernel 8.2 yielded the most satisfying result.

## 8.6 Performance Results in ImageCLEF

### 8.6.1 ImageCLEF 2009 Robot Vision Task

ImageCLEF hosted for the first time a Robot Vision task in 2009 [40, 52, 57]. The task addressed the problem of topological localization of a mobile robot using visual information.

The algorithm should have been able to provide information about the location of the robot separately for each test image (obligatory task). Training data consisted of an image sequence recorded in a five room subsection of an indoor environment under fixed illumination conditions and at a given time. Detailed information about our approach can be found in [18, 19].

The results for each run submitted to the obligatory and optional tasks are presented below. In order to encourage participation, there was no limit to the number of runs that each group could submit.

We qualified on **second** place of seven registered research groups.

| # | Group | Score |
|---|-------|-------|
| 1 | Idiap Research Institute, Martigny, Switzerland | 793.0 |
| 2 | Faculty of Computer Science, Alexandru Ioan Cuza University (UAIC), Iaşi, Romania | 787.0 |
| 3 | Computer Vision & Image Understanding Department (CVIU), Institute for Infocomm Research, Singapore | 784.0 |
| 4 | Laboratoire des Sciences de l'Information et des Systèmes (LSIS), France | 544.0 |
| 5 | Intelligent Systems and Data Mining Group (SIMD), University of Castilla-La Mancha, Albacete, Spain | 511.0 |
| 6 | Multimedia Information Modeling and Retrieval Group (MRIM), Laboratoire d'Informatique de Grenoble, France | 456.5 |
| 7 | Multimedia Information Retrieval Group, University of Glasgow, United Kingdom | 456.5 |

Table 8.4: ImageCLEF 2009 Robot Vision final results

### 8.6.2 ImageCLEF 2012 Robot Vision Task

The fourth edition of the RobotVision challenge focused on the problem of multi-modal place classification. We had to classify functional areas on the basis of image sequences, captured by a perspective camera and a kinect mounted on a mobile robot within an office environment with nine rooms. The office environment, unlike other editions, consisted in three different image sequences, which meant a large dataset represented in Table 8.1.

We had available visual (RGB) images and depth images generated from 3D cloud points. The test sequence was acquired within the same building and floor and had great variations in the lighting conditions (there have been lots of darkend sections of the offices, also, changes in environmental setting, and the acquisition procedure (clockwise and counter clockwise). Unlike in 2009, the limit to the number of runs that each group could submit was three. It can be seen that in Table 8.5 there is a group with a negative score. This is the result of the fact that the computation of the final score takes in consideration the wrong answer by descreasing it with 1.

We qualified on **third** of seven registered groups.

| # | Group | Score |
|---|---|---|
| 1 | CIII UTN FRC, Universidad Tecnolgica Nacional, Ciudad Universitaria, Cordoba, Argentina | 2071.0 |
| 2 | NUDT, Department of Automatic Control, College of Mechatronics and Automation, National University of Defense Technology, China | 1817.0 |
| 3 | Faculty of Computer Science, Alexandru Ioan Cuza University (UAIC), Iaşi, Romania | 1348.0 |
| 4 | USUroom409, Yekaterinburg, Russian Federation | 1225.0 |
| 5 | SKB Kontur Labs, Yekaterinburg, Russian Federation | 1028.0 |
| 6 | CBIRITU, Istanbul Technical University, Turkey | 551.0 |
| 7 | Intelligent Systems and Data Mining Group (SIMD), University of Castilla-La Mancha, Albacete, Spain | 462.0 |
| 8 | BuffaloVision, University at Buffalo, NY, United States | -70.0 |

Table 8.5: ImageCLEF 2012 Robot Vision final results

# Chapter 9

# Discussion

Our approach on topological localization is currently applied on an office environment of nine sections: *Corridor, ProfessorOffice, StudentOffice, LoungeArea, PrinterRoom, Toilet, VisioConference, ElevatorArea* and *TechnicalRoom*. To address the problem of recognizing these sections separately, we approached the classification with specific thresholds in taking the final decision over the selected room generated similar to the method explained here [22]. These thresholds have dependings and the created constraints have to be loosend in order to solve a more accurate result in treating situations of great similarity between two different rooms. As an example, note that one of the main inconvenient that can appear in this case is that the rooms are very connected and difficult situations can rise as the robot moves around the office. For example, if the robot is in the *Corridor*, it looks to its right and sees the *LoungeArea* but its position is still in the *Corridor*. This type of situation creates noise that cannot be neglected, therefore a proper threshold needs to treat these results that correspond to a humanized interaction with the medium. The threshold on the final decision quality was chosen to avoid erroneous localizations, thus favoring a result that doesn't specify any room and giving less correct localizations but also, less false assumptions.

This problem could also be solved with the knowledge about coordinates and temporal information. Note that we chose the experimental setup particularly difficult, without any assumption on the robot position and temporal coherency which is usually a key factor in the result quality. This means that he could be in every image in a different room, in a different time. Moreover, in a different context, using our localization system and accumulating evidences while the robot moves would obviously improve the performances.

A discussion is implied in this direction and also future research for creating a high-level processing with a level of coherence of decision and for the addition of an extension for an entailment relation.

# Chapter 10

# Conclusions and Future Work

In this work, we approach the task of topological localization without using a temporal continuity of the sequences of images using a broad variety of features for image recognition. The provided information about the environment is contained in images taken with a perspective color camera mounted on a robot platform and it represents a know office environment dataset offered by ImageCLEF.

A large scale of global and local invariant features of images was presented, investigated, and experimentally evaluated. This work gives a review of features proposed for scene recognition and refines several of them. An emphasis was placed on proposing new configurations for approaching topological localization. To analyze the features various dissimilarity measures were implemented and tested, as different features require different comparison methods. We also gave a broad overview of different comparison measures for the different types of features.

The main contribution of this work stays in quantifiable examinations of a wide variety of different configurations for a computer vision-based system and different significant results. We measured the performance of different configurations with appropriate performance measures. This implied the usage of Precision/Recall graphs and confusion matrices showed the best yelded combinations of choices towards an accurate and effective system for robot vision tasks.

*Bag-of-Visual-Words* modeling is an effective image representation in the classification task, but various representation choices like its dimension, clustering decision and feature selection have been chosen. In this paper, we have applied this technique with different parameters, approaching the problem of topological localization of a robot in an office, with no temporal and position information.

The experiments show that the configurations from different feature descriptors and distance measures depends on the proper combinations. One important aspect is to use a selection of features accounting for the different properties of the images as there is no feature capable of covering all aspects of an image.

The experiments showed the following features are suitable:

- RGB & HSV color histograms

- SIFT (Scale Invariant Feature Transform) as visual words with an Euclidean 100-means

The experiments showed also that the following image matching settings are suitable:

- RGB color histograms with Jeffrey Divergence distance & HSV color histograms with Bhattacharyya distance

- SIFT (Scale Invariant Feature Transform) matched with Minkowski distance

The experiments also proved a considerable improvement by adding a supervised learning method, SVM with RBF kernel. The results are satisfying considering future several experiments. A first starting point for this is given in this work, but further research, better and deepend analysis and relevance criteria still have to be found.

We propose introducing a feature selector and also, a robust analysis using different size of the visual word vocabularies, an expansion of appearance based features, the addition of temporal continuity which could become a great advantage if we minimize the constraints. We aim at a better understanding of the images by also adding the use of images acquired with a kinect depth sensor.

From the fact that most of the works cited are from the last couple of years, topological localization is a new and active area of research. which is increasingly producing interest and enforces further development. A first starting point for this field is given in this thesis, along with notable experimental results, but there is still room for improvement and further research.

# Appendix A

# List of Published Papers

- 2012

    - *To appear:* **Boroş E.** Towards an Accurate Topological Localization using a Bag-of-SIFT-Visual-Words Model. *IEEE 9th International Conference on Intelligent Computer Communication and Processing.* Cluj-Napoca, România, August 3 -September 1, 2012.

        *Topological localization is a problem in mobile robotics that implies the ability of an agent to self locate in an environment. In this paper, we approach the task of topological localization without using a temporal continuity of the images of the places the robot has been. The environment is represented by an office under different illumination settings acquired with a perspective camera mounted on a robot platform. We create visual vocabularies based on invariant local features and different distance-based K-means clustering. The experimental setup is performed with an One-versus-All classifier with different kernel functions that achieved success.*

- 2011

    - **Boroş E.**, Gînscă A. L., Iftene A. UAIC's Participation at Wikipedia Retrieval @ ImageCLEF 2011. *Notebook Paper for the CLEF 2011 LABs Workshop*, ISBN 978-88- 904810-1-7, ISSN 2038-4726, Amsterdam, Netherlands, September 19-22, 2011.

        *This paper describes the participation of UAIC team at the ImageCLEF 2011 competition, Wikipedia Retrieval task. The aim of the task was to investigate retrieval approaches in the context of a large and heterogeneous collection of images and their noisy text annotations. We submitted a total of six runs, focusing our effort along the textual retrieval, query expansion on English language, combined with feature extraction (Color and Edge Directionality Descriptor, CEDD). Our intention was to build a CBIR (Content-based image retrieval) system that relies on a fast indexing and retrieval practice based not only on the textual multilingual metadata, but also on the images features. The results were satisfying in the multilingual mixed search (text and images) and query expansion approach.*

- Gînscă A. L., **Boroş E.**, Metzak A., Florea A. PETI Patient Eye Tracking Interface. In *Proceedings of the 8th National Conference on Human-Computer Interaction RoCHI* 2011, ISSN 1843-4460, Pages 107-110. Bucharest, Romania, September 8-9, 2011.

  *With many communication problems faced by people with various motor and speech impediments, we propose a method of using a mobile device, based solely on eye movement. This article describes PETI system (Patient Eye Tracking Interface), which helps patients, giving them a means of interaction with medical staff, but also to connect with online environment.*

- 2010

  - **Boroş, E.**, Roşca, G., Iftene, A. Using SIFT Method for Global Topological Localization for Indoor Environments. In *C. Peters et al. (Eds.): CLEF 2009, LNCS* 6242, Part II (Multilingual Information Access Evaluation Vol. II Multimedia Experiments). Pages 277-282. ISBN: 978-3-642-15750-9. Springer, Heidelberg, 2010.

    *The paper represents a brief description of our system as one of the solutions to the problem of global topological localization for indoor environments. The experiment involves analyzing images acquired with a perspective camera mounted on a robot platform and applying a feature-based method (SIFT) and two main systems in order to search and classify the given images. To obtain acceptable results and improved performance improvement, the algorithm acquires two main maturity levels: one capable of running in real-time and taking care of the computers resources and the other one capable of classifying correctly the input images. One of the principal benefits of the developed system is a server-client architecture that brings efficiency to the table along with statistical methods that improve the quality of data with their design.*

- 2009

  - **Boroş E.**, Roşca G., and Iftene A. Uaic: Participation in Image-CLEF 2009 RobotVision task. Proceedings of the CLEF 2009 Workshop. Corfu, Greece, September 30-October 2, 2009.

    *This year marked our first participation at the RobotVision task a new task from ImageCLEF competition. The paper represents a brief description of our system as the solution to the problem of topological localization of a mobile robot using visual information. We were asked to determine the topological location of a robot based on images acquired with a perspective camera mounted on a robot platform. And so, we decided that we don't need an incremental learning system and*

*we approached a statistical method that always will work the best results. We used to apply a feature-based method (SIFT) and two main systems in order to search and classify the given images written by us. At the same time, the systems preserve the recognition performance of the batch algorithm. The first system is reordering the images so we can get the most important/representative images for a room's category. This is done using SIFT. The second system is a brute one, just for testing the differences between this one and the first one, not selecting the representative images. We acquired a separation in directories for every room capturing the key points saved in files for every image (or representative) from every room. About the changes in the environment, the SIFT algorithm occupies itself. The entire recognition system consists in a server client architecture. Server is processing one single search at a time, and after the search ends connection with the client closes. The resulting file is the asked-for file with the final results for the batched images.*

– **Boroş E.**, Roşca G., and Iftene A. Using sift method for global topological localization for indoor environments. C. Peters et al. (Eds.): CLEF 2009, LNCS, 6242:277–282, 2009.

*The paper represents a brief description of our system as one of the solutions to the problem of global topological localization for indoor environments. The experiment involves analyzing images acquired with a perspective camera mounted on a robot platform and applying a feature-based method (SIFT) and two main systems in order to search and classify the given images. To obtain acceptable results and improved performance improvement, the algorithm acquires two main maturity levels: one capable of running in real-time and taking care of the computers' resources and the other one capable of classifying correctly the input images. One of the principal benefits of the developed system is a server-client architecture that brings efficiency to the table along with statistical methods that improve the quality of data with their design.*